\documentclass[conference]{IEEEtran}
\usepackage{amsmath,amsfonts}
\usepackage{multirow}
\usepackage{array}
\usepackage{cite}

\usepackage{algpseudocode}
\usepackage{amsmath}
\usepackage{graphicx}
\usepackage{booktabs} 
\usepackage[caption=false,font=normalsize,labelfont=sf,textfont=sf]{subfig}
\usepackage{hyperref} 
\usepackage{textcomp}
\usepackage{stfloats}
\usepackage{url}
\usepackage{verbatim}
\usepackage{graphicx}
\usepackage{cite}
\usepackage{subcaption}
\begin{document}

\title{ClutterNav: Gradient-Guided Search for Efficient 3D Clutter Removal with Learned Costmaps}

\author{\IEEEauthorblockN{Navin~Sriram~Ravie, Keerthi~Vasan~M, Bijo~Sebastian}
\IEEEauthorblockA{\textit{Department of Engineering Design} \\
\textit{Indian Institute of Technology, Madras}\\
Chennai, India \\
bijo.sebastian@iitm.ac.in}
}

\maketitle

\begin{abstract}
Dense clutter removal for target object retrieval presents a challenging problem, especially when targets are embedded deep within densely-packed configurations. It requires foresight to minimize overall changes to the clutter configuration while accessing target objects, avoiding stack destabilization and reducing the number of object removals required. Rule-based planners when applied to this problem, rely on rigid heuristics, leading to high computational overhead. End-to-end reinforcement learning approaches struggle with interpretability and generalizability over different conditions. To address these issues, we present ClutterNav, a novel decision-making framework that can identify the next best object to be removed so as to access a target object in a given clutter, while minimising stack disturbances. ClutterNav formulates the problem as a continuous reinforcement learning task, where each object removal dynamically updates the understanding of the scene. A removability critic, trained from demonstrations, estimates the cost of removing any given object based on geometric and spatial features. This learned cost is complemented by integrated gradients that assess how the presence or removal of surrounding objects influences the accessibility of the target. By dynamically prioritizing actions that balance immediate removability against long-term target exposure, ClutterNav achieves near human-like strategic sequencing, without predefined heuristics. The proposed approach is validated extensively in simulation and over real-world experiments. The results demonstrate real-time, occlusion-aware decision-making in partially observable environments.

\end{abstract}

\section{Introduction}
\begin{figure}[t]
    \centering
    \includegraphics[width=0.8\linewidth]{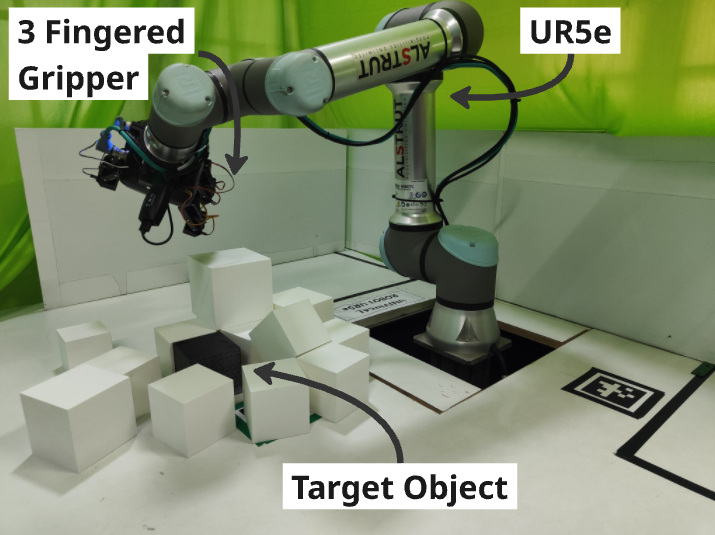}
    \caption{Hardware setup used for experimental validation.}
    \label{fig:revintro}
\end{figure}

Robotic bin picking \cite{binpick}, a challenging benchmark in industrial automation, has seen continuous research in recent years. Decomposed as a sequence of simple pick-and-place operations in traditional manufacturing lines, the planning is often straightforward as they avoid clutter through the use of predefined fixtures and sequenced workflows. On the other hand, most real-world scenarios involve targets buried within dense, interconnected clutter, where accessing one object affects the stability and positioning of multiple others as shown in Fig. ~\ref{fig:revintro}, hereafter referred to as \textbf{Dense Clutter Removal}.

Traditionally, bin picking assumes direct accessibility to target objects. However, in the case of Dense Clutter Removal systematic removal of neighboring objects while minimizing overall disturbance to the clutter configuration is needed.  Unlike conventional bin-picking that focuses on individual object retrieval, Dense Clutter Removal requires strategic sequencing that preserves structural stability of the remaining clutter while creating access to deeply buried targets in minimum steps. In real-world settings an object inside the clutter could be disturbed due to collision with the robotic manipulator or with another object being removed by the manipulator. It could also be disturbed due to structural instability in the clutter caused due to the removal of another object. These interactions depend on complex rigid-body interactions involving multiple objects, which are often difficult to model with closed-form mathematical equations. 

This problem manifests in diverse domains such as pharmaceutical packaging, e-commerce fulfillment where items are densely packed in bins, and disaster response where survivors or supplies may be trapped beneath debris. Each scenario demands disturbance-minimizing strategies that current bin-picking approaches fail to address.

Early solutions in this direction relied on rule-based sequencing \cite{{old}}, where grasp candidates are prioritised using geometric heuristics like proximity or vertical accessibility. While effective for semi-structured bins with predefined and observable features, these methods fail in the presence of dense, unordered clutter. Hierarchical task-motion planning (TAMP) \cite{tamp}, with interleaved symbolic task planning, was developed to address such problems. Frameworks like Nam et al.’s \cite{nam} occlusion-minimising planner addressed the issues partially by modelling object inter-dependencies as a directed graph. But TAMP is computationally expensive, and its application is currently limited to 2D target retrieval in structured clutter, with little correlation and emphasis on the Dense Clutter Removal problem.  

\begin{figure}[t]
    \centering
    \includegraphics[width=0.85\linewidth]{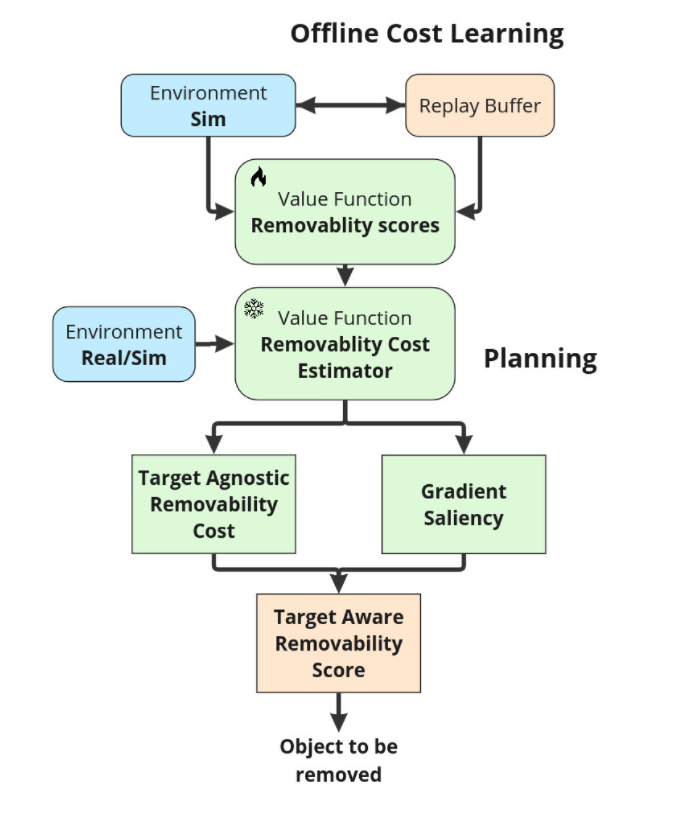}
    \caption{Overview of the proposed ClutterNav framework. (Top) Real-world setup showing a UR5e arm with a 3F140 gripper executing clutter removal to retrieve a target object. (Bottom) System pipeline combining offline removability cost learning from simulation and online gradient-guided planning for efficient and safe target retrieval.}
    \label{fig:graphabs}
\end{figure}

Recent developments in machine learning, particularly in Reinforcement Learning (RL)  have made decision-making and planning possible, even based on raw unstructured point cloud inputs. While such an end-to-end policy would seem favourable, generating such policies requires careful tuning of the reward landscape, which becomes a challenging task in itself, rewarding disturbance minimization and target removal at the same time. Interpretability of such policies limit real-world application due to their black-box nature \cite{rlpick1}\cite{rlpick2}, and such policies trained in simulation may struggle to effectively perform in real-world environments. The use of learnt encoders needs careful data augmentation and randomization during training to reduce the sim2real gap. We show that first learning removability costs for efficient removal and then steering the removal towards the target via implicit goal conditioning without retraining produces satisfactory results.

This work addresses the fundamental challenge of Dense Clutter Removal - strategically removing objects to retrieve targets while minimising overall configuration disturbance. We present ClutterNav, a novel framework that combines learned cost estimation with gradient-guided decision making for efficient dense clutter removal. Our key contributions are:
\begin{itemize}
    \item A \textbf{lightweight decision-making framework} that formulates dense clutter removal as a sequential object removal problem, combining learned removability costs with real-time gradient-guided planning for minimal-disturbance target retrieval.
    \item A \textbf{goal-agnostic removability cost estimator} trained through demonstrations using a critic-centric reinforcement learning approach that predicts object removal costs without explicit target conditioning.
    \item An integrated gradients approach for \textbf{implicit goal conditioning} that leverages gradient saliency analysis to steer object removals toward target exposure, providing interpretable decision sequences without predefined heuristics.
    \item We perform extensive experimental validation, showing the efficacy of the proposed approach in tackling the Dense Clutter Removal problem.
\end{itemize}



\section{Methodology}
The \textbf{Dense Clutter Removal} problem, as shown in Fig.~\ref{fig:graphabs}, can be viewed as jointly minimizing two conflicting objectives: minimum number of object retrievals to reach the target and reducing the disturbance to the overall clutter while  doing so. In extreme cases, the minimum number of object removals is one, which is directly picking up the object, but such an action may destabilize the surrounding clutter. Conversely, if disturbance minimization is prioritized, the target object may be retrieved much later. We solve this high-dimensional decision-making problem by learning the cost to remove an object and steering the search with gradients. The cost for picking any object within a given scene is predicted using a goal-agnostic model learnt through demonstrations. To prioritise the removal of objects blocking the target, we use the learnt model for predicting safe actions and utilise the gradients of this learnt model to steer object removals around the target object, providing implicit goal conditioning. The proposed approach is validated for cuboidal objects as well as other standard objects from the YCB object dataset\cite{ycb}.

While the proposed method does not guarantee global optimality or theoretical bounds on convergence, it represents one of the first attempts to systematically address the problem in its current form. Despite this, it demonstrates strong empirical performance and offers a promising foundation for future exploration in clutter-aware decision-making. It should be noted that we do not propose any novel methodologies for grasp planning since prior literature \cite{gpd, quickgrasp} has reliably demonstrated object grasping with partial point cloud inputs.

\subsection{Preliminaries}

\textbf{Disturbance Quantification}: We define \textit{disturbance} as the cumulative spatial displacement of objects in the clutter configuration caused by a removal action. Specifically, disturbance $D_{\text{total}}$ is computed as the sum of Euclidean distances between object centroids before and after each removal action:

\begin{equation}
D_{\text{total}} = \sum_{i \in \mathcal{O}_{\text{remaining}}} \| \mathbf{c}_i^{\text{post}} - \mathbf{c}_i^{\text{pre}} \|_2
\end{equation}

where $\mathbf{c}_i^{\text{pre}}$ and $\mathbf{c}_i^{\text{post}}$ are the centroid positions of object $i$ before and after the action, respectively, and $\mathcal{O}_{\text{remaining}}$ represents all objects except the one being removed. 

\textbf{Removability:} The ease and safety of removing an object without destabilizing the clutter configuration, quantified as the inverse of the learnt removability cost.

\textbf{MDP Formulation:} We model object removal as an MDP $(\mathcal{S}, \mathcal{A}, \mathcal{P}, \mathcal{R})$ where $\mathcal{S}$ represents clutter configurations, $\mathcal{A}$ represents object removal actions, $\mathcal{P}$ captures transition dynamics, and $\mathcal{R}$ encodes disturbance minimization objectives.

\textbf{State Representation:} The state consists of normalised geometric and spatial features for all observable objects, represented as a padded feature matrix to accommodate varying object counts.

\subsection{Simulation Environment and Clutter Generation}

\begin{figure}[t]
    \centering
    \includegraphics[width=0.8\linewidth]{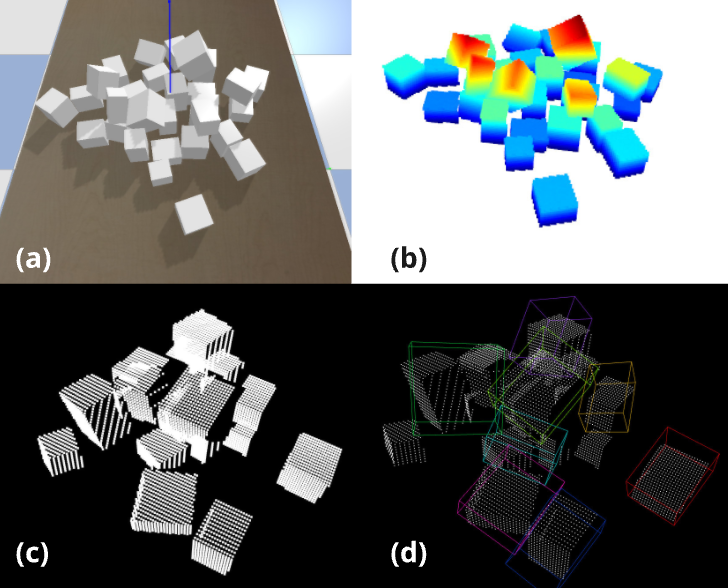}
    \caption{Sim Setup, (a) Clutter on the table (b) Raw point cloud (c) Observable Point Cloud (d) Cuboid fitting on point cloud}
    \label{fig:sim}
\end{figure}
A high-fidelity simulation environment was set up in PyBullet to learn the cost map and validate it towards object manipulation in cluttered 3D spaces. PyBullet’s physics engine replicates physical interaction between the objects and object displacement in simulation, mirroring real-world interactions. The cost-map learning was done using cuboid objects as shown in Fig.~\ref{fig:sim}, with randomised dimensions in the range of $70mm - 90 mm$, and a constant density of $140kg/m^3$.

Realistic clutter was generated in simulation using a structured, layered cuboid spawning mechanism. Cuboids are randomly dropped onto a planar surface in multiple layers, allowing physics-based settling. This creates cluttered configurations where objects may rest atop others, leading to occlusions and varied object dependencies, as shown in Fig.~\ref{fig:sim}. 

To simulate realistic interactions, each object is assigned frictional properties that affect stability and responses to external disturbances. During the cost-map learning phase, objects are removed by deleting them from the simulated scene.

\subsection{Instance Segmentation on Point Clouds}

In order to ensure easy translation to real-world applications a perception pipeline is developed to estimate individual cuboid pose from partial point cloud data. The proposed perception pipeline uses a three-stage geometric approach. To begin with, surface normals are estimated using Principal Component Analysis (PCA) on local point neighbourhoods. 
After normal estimation, curvature-constrained region growing is applied to segment planar surfaces. This allows for iteratively expanding clusters by merging points based on angular deviation and curvature patterns for the normals. Finally, axis-aligned bounding boxes (AABB) are fitted to each cuboid candidate to extract its dimensions $(L, W, H)$, centroid $C$, and orientation. The raw observed point cloud and the AABB fit on it are shown in Fig. ~\ref{fig:sim}. 

\begin{figure}[t]
    \centering
    \includegraphics[width=0.9\linewidth]{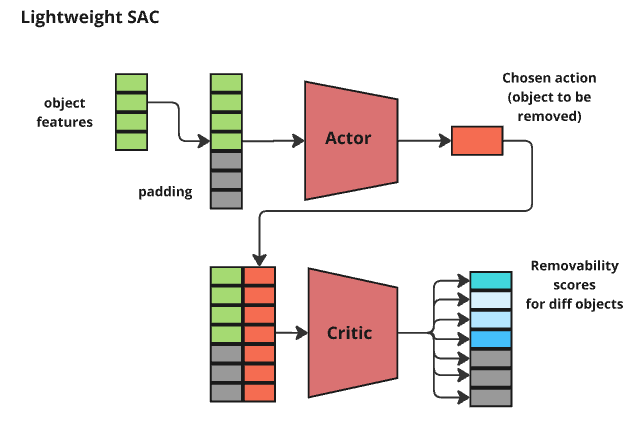}
    \caption{Lightweight SAC Architecture displaying the observable object features (green), passed through the Actor after padding (grey), outputting the action (orange) concatenated with the padded feature vector and passed through the critic, giving the removability scores for the observable objects }
    \label{fig:sac}
\end{figure}

It should be noted that latest SOTA approaches can also be used by employing SAM \cite{SAM} for extracting object level segmentation masks, and using the depth and intrinsics for getting the object level point cloud. Depth based segmentation was preferred over vision based segmentation strategy of SAM to distinguish between cuboids of similar color as used the experimental validation. While our perception pipeline demonstrates AABB fitting for cuboidal objects, the extracted geometric features can generalise to arbitrary object shapes. The core feature extraction approach outlined in Section II.D applies universally to any segmented object regardless of geometry.

\subsection{Reinforcement Learning for Cost Estimation}
\subsubsection{\textbf{Feature Extraction} }
A feature-driven reinforcement learning framework is proposed to estimate the removability score for objects in cluttered 3D environments. Geometric features that generalize across object types are used, computed from instance segmentation results to ensure consistent behavior across diverse object geometries. All features were min-max normalised per scene for consistency. The features can be classified into two categories: 
\begin{itemize}
    \item Spatial context: Each object’s $(x,y,z)$ position, orientation, vertical offset, and $L2$ distance to the clutter centroid was used to quantify spatial distribution and aid grasp selection.  
    \item Geometric properties: Bounding box volume was used to characterise object shape, while KD-tree-based neighbour distances were used to estimate isolation and occlusion.  
\end{itemize}

Unlike deep learning methods such as PointNet++ \cite{pn++}, the proposed method uses known features and a Soft Actor-Critic (SAC) \cite{sac} framework to improve interpretability and efficiency. SAC enables exploration of risky removals, avoids overfitting to suboptimal grasps, and increases sample efficiency through off-policy learning in the sparse-reward setting. 

\subsubsection{\textbf{Why Handcrafting?} }Deterministic encoding ensures identical outputs for identical geometric configurations, enhancing repeatability, a critical limitation of learned encoders that exhibit variability due to training stochasticity. Domain knowledge integration focuses learning on physically meaningful relationships, avoiding spurious correlations in out-of-distribution scenarios where learned features often fail.  Recent literature demonstrates that learnt 3D encoders suffer from sensitivity to geometric perturbations, occlusions, and domain shifts, requiring extensive data augmentation for robust generalisation \cite{dgmvp}\cite{partGen}. Although learned encoders would seem attractive, they do not generalize across different cameras, camera poses and other uncontrollable factors. Simple and consistent handcrafted features perform better as demonstrated in this work. 

\subsubsection{\textbf{Removability Critic} }Instead of focusing on policy optimisation, a critic-centric design where the Q-network serves as a differentiable cost estimator was used to capture object dependencies. To demonstrate the efficacy of the approach, a relatively simple and lightweight sequential Multi-Layer Perceptron (MLP) with three linear layers to represent the Critic Network was used, as shown in Fig.~\ref{fig:sac}. Each object in the scene is treated as a discrete action (remove/keep), with the critic evaluating all objects in parallel through Q-value queries. To handle varying numbers of objects per scene, a dynamic state padding mechanism was used, where removed objects are masked with zero vectors, preventing the need for retraining across different clutter configurations. While it would be convenient to use state-of-the-art algorithms like transformers and GNN architectures, a standard MLP was used to demonstrate that the cost function could be learnt efficiently with $90\%$ fewer parameters.

The critic network was trained to predict the long-term feasibility of removing objects by minimizing an error function between predicted and actual rewards. Unlike standard actor-critic models that optimize policy gradients, this setup allows the critic to independently learn removability costs, making decisions interpretable and stable across varying clutter scenarios.

\subsubsection{\textbf{Removability Scores }}
Object removal from clutter is defined as a Markov Decision Process (MDP) where $S$ represents the state feature vector. Every row in $S$ corresponds to the features extracted from an observable object in the scene. Actions $a$ in the MDP are formulated as the object to be removed, which is represented through the row index $i$, which we call the object index, in the state feature vector.

The trained critic is used to calculate the removability score $Q_{i}$, which is the Q value corresponding to the current state feature vector $S$ along with the action $i$ corresponding to the object index to be removed.

\begin{equation} \label{eq:raw_score}
    Q_i = Q(s, i)   
\end{equation}

\subsubsection{\textbf{Reward Function Design}} The reward function was structured to penalise object disturbances:
\begin{equation}
r(s_t, a_t) = - \mathcal{O}_d - \gamma \cdot D_{\text{total}}
\end{equation}
\begin{itemize}
    \item $\mathcal{O}_d$ is the number of objects moved
    \item $D_\text{total}$ is the total disturbance in metres
    \item $a_t$ is the object index for the object removed
\end{itemize}
We choose $\gamma=0.1$ but the performance is not strictly dependent on any $\gamma<1$, for the above reward. This formulation ensures that actions minimising unintended movements are favoured, reinforcing stability-aware removability learning.


\subsection{Gradient Guided Object Importance Estimation}
\label{sec:gradient_guidance}

To prioritise objects that maximally expose the target while minimising disturbances, we compute gradient-based importance scores. These scores quantify how removing an object affects the target’s accessibility. The proposed method combines learned cost predictions with explainable gradient analysis through three key innovations: (1) Exact Influence Gradient Formulation, (2) A path-integrated attribution for stable importance estimates, and (3) spatial attention through physical proximity.

\subsubsection{Exact Influence Gradient Formulation}
The first-order gradient of the target object's predicted removability score $Q(s,a_{\text{target}})$ with respect to each object's state features is given by:

\begin{equation} \label{eq:raw_gradient}
    \nabla_i Q = \frac{\partial Q(s, i_{\text{target}})}{\partial s_i} \in \mathbb{R}^d
\end{equation}
Here  $i_\text{target}$ is the object index of the target in the state feature vector.
This raw gradient $\nabla_i Q$ indicates how would slightly perturbing object $i$'s state features could affect the target objects removal cost. Large gradient magnitudes indicate objects which significantly impacts the target's removability.

\subsubsection{Integrated Gradients} \label{sec:ig}
To obtain stable importance estimates, we compute path-integrated gradients \cite{ig}, a common approach used in explainability to attribute a model's output to input features. Traditional gradient-based attribution suffers from saturation effects and local sensitivity. The proposed path-integrated gradients address this by quantifying counterfactual impacts through baseline-to-input interpolation, revealing how removing an object (vs perturbing it) affects target accessibility via:
\[
\mathcal{G}_i= \int_{0}^{1} \frac{\partial Q}{\partial s_i} \bigg|_{s_{\alpha}} d\alpha,
\]

\begin{figure}[t]
    \centering
    \includegraphics[width=0.7\linewidth]{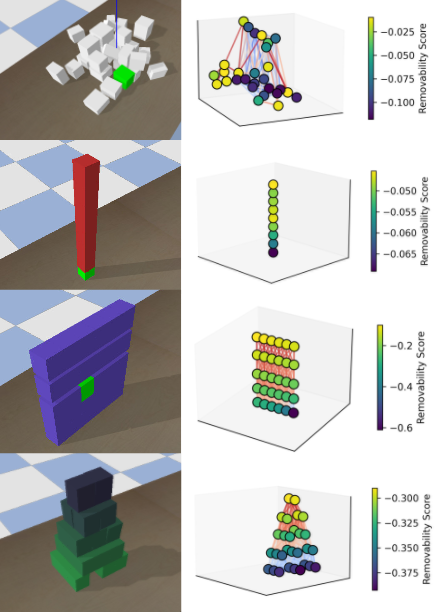}
    \caption{Cost Landscape for our experimental arrangements - top to bottom - Random Clutter, Stack, Wall and Pyramid}
    \label{fig:1}
\end{figure}

This provides physics-grounded explanation as the gradient magnitudes directly correlates with an object's role in occlusion chains.

\begin{equation} \label{eq:ig}
    \mathcal{G}_i = \underbrace{\exp(-\lambda \|p_i - p_t\|_2)}_{\text{Spatial decay}} \cdot \frac{1}{K}\sum_{k=1}^K \left[ \nabla_i Q(s^{(i)}_k) \cdot (s_i - s^{(i)}_{\text{base}}) \right]
\end{equation}

Where:
\begin{itemize}
 \item $p_i$ and $p_t$ are the positions of the $i^{th}$ object and the target object respectively
    \item $s^{(i)}_{\text{base}}$ is the baseline state with object $i$ zeroed out
    \item $s^{(i)}_k = s^{(i)}_{\text{base}} + \alpha_k(s - s^{(i)}_{\text{base}})$ for $\alpha_k \sim U(0,1)$
    \item $\lambda$ controls spatial attention ($\lambda=2$ in implementation)
    \item $K=5$ Monte Carlo samples
    \item Here, $K=5$ determines the numerical precision of the integrated gradients approximation (trading accuracy for computational efficiency), while $\lambda=2$ controls spatial attention by exponentially weighting objects based on their proximity to the target, providing a smoothening filter to the extract gradients.
    
\end{itemize}




\subsubsection{Complete Importance Score}
The final importance score combines normalised cost predictions with gradient magnitudes:

\begin{equation} \label{eq:final_score}
    \mathcal{S}_i = \underbrace{\frac{Q_i - Q_{\text{min}}}{Q_{\text{max}} - Q_{\text{min}}}}_{\text{Normalized cost}} + \underbrace{\frac{\|\mathcal{G}_i\| - \mathcal{G}_{\text{min}}}{\mathcal{G}_{\text{max}} - \mathcal{G}_{\text{min}}}}_{\text{Normalized influence}}
\end{equation}


The ratio was tuned to be  1:1 to give best results through parameter sweeps. This balanced approach enables simultaneous disturbance minimisation and target retrieval, as demonstrated in the next Section.

\section{Validation over simulation}

The trained object removal cost estimator framework was evaluated over simulation,  on structured clutter configurations not encountered during learning. These arrangements, depicted in Fig. \ref{fig:1}, encompass:

\begin{itemize}
    \item \textbf{Random Clutter} – A complex scene generated by randomly dropping over 30 cuboids onto a flat surface.
    \item \textbf{Vertical Stack} – A structured column of 10 cuboids.
    \item \textbf{Wall Formation} – A densely packed arrangement of 30 cuboids, with the target object embedded randomly.
    \item \textbf{Pyramidal Structure} – A stable formation of 30 cuboids resembling a pyramid.
\end{itemize}


Given the novelty of the proposed approach, direct comparisons with rule-based heuristics or Task and Motion Planning (TAMP) baselines are impractical due to fundamental differences in problem formulation and state-space complexity. Instead, the proposed approach was benchmarked against the following strategies to demonstrate efficacy.

\begin{figure}[t]
    \centering
    \includegraphics[width=0.84\linewidth]{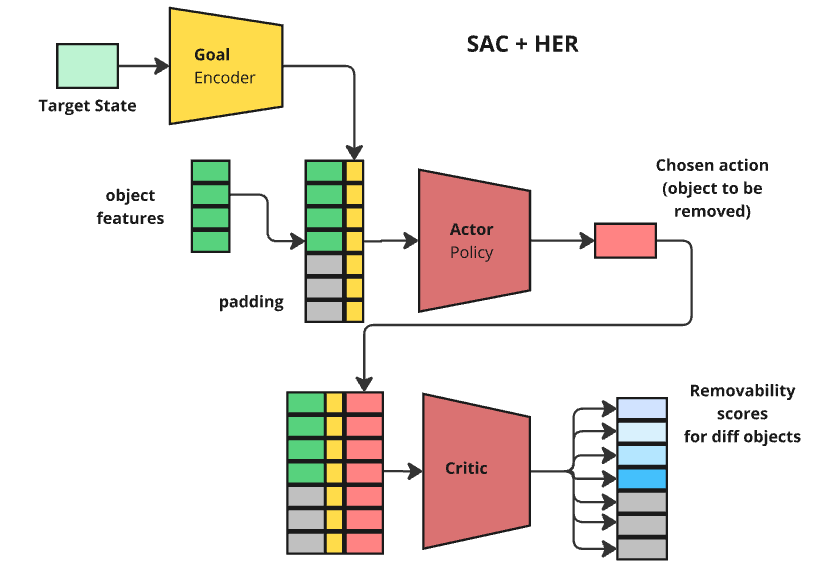}
    \caption{SAC+HER framework using a learnt goal feature encoder to provide goal conditioning using learnt features along with the Actor and Critic using the standard SAC loss}
    \label{fig:sac_her}
\end{figure}

\begin{table}[b]
    \centering
    \caption{Comparison of disturbance of the COM using ground truth from Pybullet (in meters) across different policies over 100 retrieval trials}
    \label{tab:disturbance}
    \resizebox{\columnwidth}{!}{ 
    \begin{tabular}{lcccc}
        \toprule
        \textbf{Scenario} & \textbf{Random Policy} & \textbf{Conservative} & \textbf{Ours} & \textbf{SAC+HER} \\
        \midrule
        Clutter (30 objects) & $0.27 \pm 0.18$ & $0.05 \pm 0.08$ & \textbf{0.02 $\pm$ 0.06} & $0.36 \pm 0.22$ \\
        Pyramid (30 objects) & $0.83 \pm 0.80$ & $\mathbf{0.006 \pm 0.003}$ & $0.11 \pm 0.15$ & $0.83 \pm 0.75$ \\
        Stack (10 objects) & $1.04 \pm 0.53$ & $0.08 \pm 0.03$ & $\mathbf{0.08 \pm 0.03}$ & $1.01 \pm 0.56$ \\
        Wall (30 objects) & $1.26 \pm 0.67$ & $\mathbf{0.037 \pm 0.012}$ & $0.13 \pm 0.16$ & $1.52 \pm 0.81$ \\
        \bottomrule
    \end{tabular}}
\end{table}

\begin{table}[b]
    \centering
    \caption{Number of object removals required to reach the target over 100 retrieval trials}
    \label{tab:removals}
    \resizebox{\columnwidth}{!}{ 
    \begin{tabular}{lcccc}
        \toprule
        \textbf{Scenario} & \textbf{Random Policy} & \textbf{Conservative} & \textbf{Ours} & \textbf{SAC+HER} \\
        \midrule
        Clutter (30 objects) & $15.63 \pm 8.93$ & $16.27 \pm 8.73$ & \textbf{8.11 $\pm$ 10.22} & $16.86 \pm 7.94$ \\
        Pyramid (30 objects) & $16.06 \pm 8.52$ & $16.99 \pm 8.36$ & \textbf{8.8 $\pm$ 11.32} & $15.25 \pm 8.86$ \\
        Stack (10 objects) & $5.84 \pm 2.72$ & $5.47 \pm 2.68$ & \textbf{5.1 $\pm$ 2.71} & $5.63 \pm 2.79$ \\
        Wall (30 objects) & $14.64 \pm 8.12$ & $15.55 \pm 8.36$ & \textbf{8.45 $\pm$ 10.29} & $14.86 \pm 8.63$ \\
        \bottomrule
    \end{tabular}}
\end{table}


\begin{itemize}
    \item \textbf{Random Policy}: This is the worst-case scenario baseline where object selection follows a uniform random policy.
    \item \textbf{Conservative Safe Policy}: A purely cost-driven selection strategy, to serve as an ablation study, that greedily removes objects based on the highest predicted removability score. The cost model was trained for 3000 steps in about 8 minutes on an RTX 3060 GPU.
    \begin{equation}
    \text{action} = \arg\max(Q_{\text{score}_i}).
    \end{equation} 
    \item \textbf{Goal-Conditioned RL Baseline:} A Soft Actor-Critic model with Hindsight Experience Replay (SAC+HER) is used as the goal-conditioned RL baseline. SAC+HER addresses sparse rewards encouraging quicker retrieval by re-labelling failed episodes as alternative successes and uses structured learnt goal encoding replacing the one-shot representation. (see Fig.~\ref{fig:sac_her}). SAC+HER was trained for over 150,000 steps on an RTX 3060 GPU. 
\end{itemize}

\subsection{Results and Inference}
The effectiveness of the proposed approach for target retrieval in cluttered settings was evaluated. The  results are discussed below:

\subsubsection{Effectiveness of Cost-Based Estimation}
As shown in Table~\ref{tab:disturbance}, we achieve minimal disturbances in all environments in the order of few centimeters. Our method reduces disturbances to 2cm in random clutter (vs. 27cm for random policies) while requiring 50 per cent fewer removals than conservative baselines     (Table~\ref{tab:removals}).   

\subsubsection{Comparison with SAC+HER}
Despite SAC+HER being a strong baseline, our approach outperforms it in both efficiency and stability since SAC+HER tries to learn a hard policy that satisfies both conflicting objectives: disturbance minimization and the number of object retrievals to reach the target. Table~\ref{tab:removals} shows that our method reaches the goal with fewer object removals and overall distortion. Although this could be surpassed with a more complex end-to-end policy with better architectures and careful hand-tuning, one-shot encoding with a simple MLP works better, as shown empirically.

\subsubsection{The Role of Conservative Policies and Goal Conditioning}
The conservative policy minimizes disturbances but at the cost of requiring excessive removals. The proposed approach retains the conservative policy’s low-disturbance characteristics while drastically reducing the number of removals, as shown in Table~\ref{tab:removals}.

\subsubsection{Random Policy as a Worst-Case Baseline}
The random policy illustrates the chaotic effects of uninformed decision-making. The contrast between the proposed method and the random baseline further validates the effectiveness of learned cost-based reasoning.

\subsubsection{ Performance on YCB objects}

\begin{table}[b]
    \centering
    \caption{Number of object (YCB)  removals required to reach the target over 50 trials }
    \label{tab:removals1}
    \resizebox{\columnwidth}{!}{ 
    \begin{tabular}{lcccc}
        \toprule
        \textbf{Scenario} & \textbf{Random Policy} & \textbf{Conservative} & \textbf{Ours} & \textbf{SAC+HER} \\
        \midrule
        Blue Moon (30 objects) & $13.18 \pm 8.35$ & $8.25 \pm 4.73$ & \textbf{4.03 $\pm$ 3.57} & $14.6 \pm 9.2$ \\
        Soap (30 objects) & $14.4 \pm7.64$ & $15.0 \pm 0$ & \textbf{7.03$\pm$2.63} & $15.3 \pm 9.21$ \\
        Hammer (30 objects) & $15.34 \pm 8.38$ & $10.32\pm 6.46$ & \textbf{ 2.68 $\pm$ 2.25 } & $  15.04\pm 7.62$ \\
        \bottomrule
    \end{tabular}}
\end{table}
        
\begin{figure}
    \centering
    \includegraphics[width=1\linewidth]{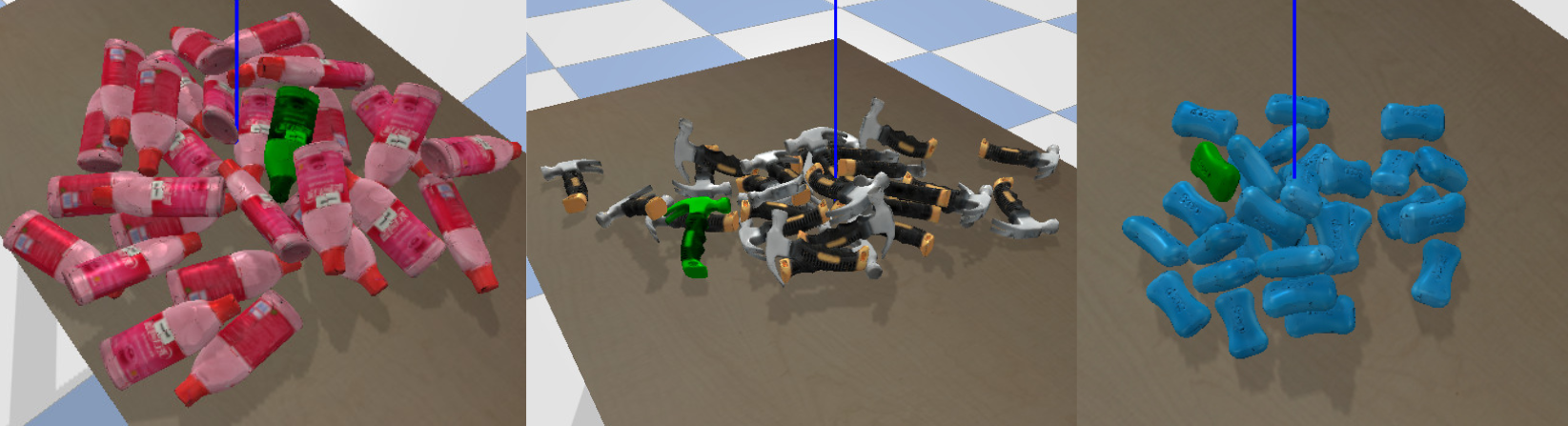}
    \caption{YCB Objects used with varying geometries 1. Blue Moon, 2. Hammer, 3. Soap.}
    \label{fig:ycb}
\end{figure}

\begin{table}[b]
    \centering
    \caption{Average disturbance (YCB) in the clutter over 50 trials}
    \label{tab:removals2}
    \resizebox{\columnwidth}{!}{ 
    \begin{tabular}{lcccc}
        \toprule
        \textbf{Scenario} & \textbf{Random Policy} & \textbf{Conservative} & \textbf{Ours} & \textbf{SAC+HER} \\
        \midrule
        Blue Moon (30 objects) & $0.37 \pm 0.22$ & $0.06 \pm 0.05$ & \textbf{0.02 $\pm$ 0.03} & $ 0.35 \pm 0.26$ \\
        Soap (30 objects) & $0.26\pm0.14$ & \textbf{0.009 $\pm$ 0.014} &  $0.021\pm 0.034$ & $0.138 \pm 0.11$ \\
        Hammer (30 objects) & $0.38 \pm 0.42$ & $0.08\pm 0.18$ & \textbf{0.008$\pm$ 0.014} & $ 0.253\pm 0.606$ \\
        \bottomrule
    \end{tabular}}
\end{table}
In order to evaluate the generalisabilty of the proposed approach we evaluated the 
cost estimator, trained on cuboidal objects, over more complex objects from the YCB object Dataset, such as blue moon, soap and the hammer as shown in Fig.~\ref{fig:ycb}. To evaluate generalization across object geometries, we applied the same feature extraction methodology used for cuboidal objects. The spatial context and geometric properties described in Section II.C were computed using each YCB object's AABB representation, demonstrating that our geometric feature set generalizes effectively to complex, near-cuboidal shapes. The results for clutter removal using the trained network is shown in Table ~\ref{tab:removals1} and ~\ref{tab:removals2}. The results indicate that the model trained transfers well to other complex shapes.

\section{Hardware Validation}
\label{sec:hardware}


We validate our proposed pipeline on 3D printed white cuboids with different dimensions forming the clutter with the target cuboid in black for easier identification. For grasping the objects, a simple pre-defined grasp primitive designed for cuboids was used. The use of clear geometric and spatial features leading to a near-zero sim2real gap has already been discussed, since it uses deterministic feature extraction from segmented pointclouds. 

The hardware validation was performed as follows. The next object to be removed was determined using the proposed algorithm. Once the object is identified, pre-grasp pose was calculated for the object and the manipulator was moved to the desired pose. A UR5e collaborative Manipulator with a ROBOTIQ three finger gripper, operating in pinch mode to perform antipodal grasps, was used. The removed objects were placed at a predefined point. 

We perform evaluation over 10 diverse clutter configurations , which can broadly be classified into three categories as shown in Table ~\ref{tab:real_world}.

\begin{figure*}[htbp]
    \centering
    \includegraphics[width=1\linewidth]{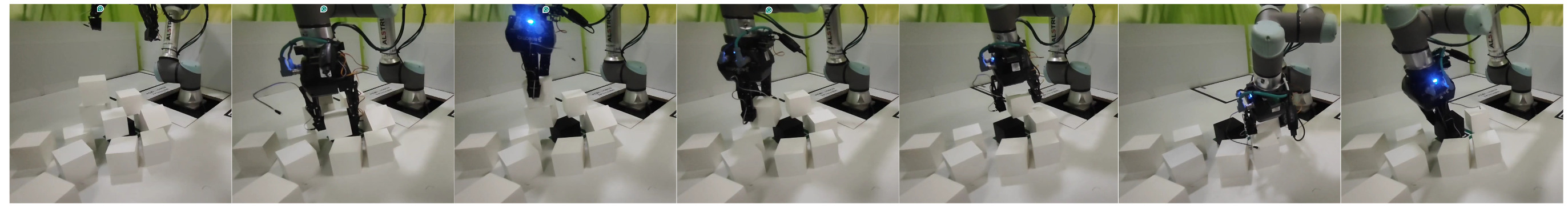}
    \caption{Real world removal sequence, target is the cuboid in black color}
        \label{fig:rws}
\end{figure*}
\begin{figure*}[htbp]
        \centering
        \includegraphics[width=1\linewidth]{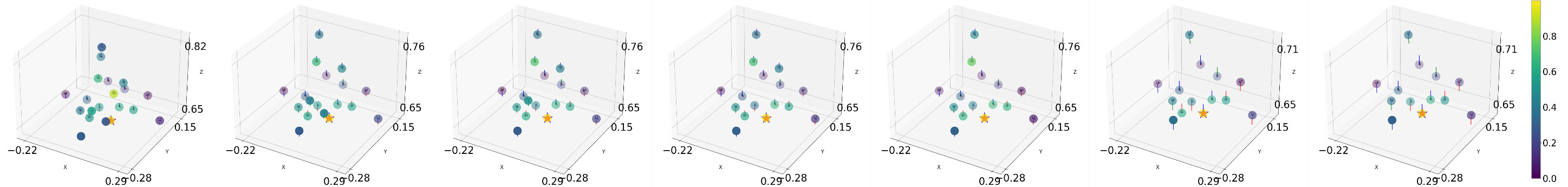}
        \caption{Gradient influence score at each step, target denoted star}
        \label{fig:rs}
    \end{figure*}
\begin{figure*}[htbp]
    \centering
    \includegraphics[width=1\linewidth]{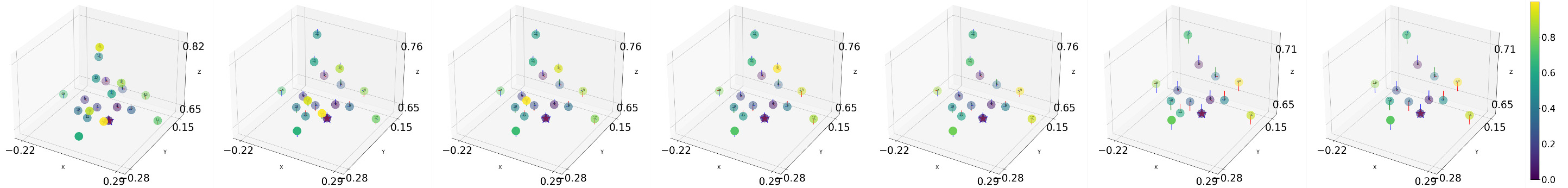}
    \caption{Removability scores at each step, target denoted star}
        \label{fig:gs}
\end{figure*}
\begin{figure*}[htbp]
        \centering
        \includegraphics[width=1\linewidth]{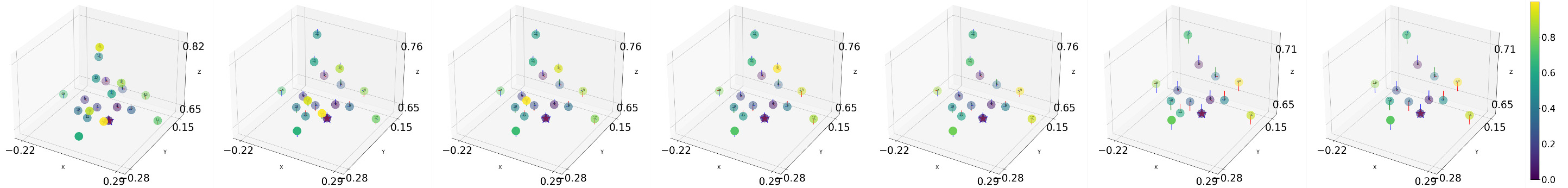}
        \caption{Combined decision scores at each step, target denoted star}
        \label{fig:cd}
    \end{figure*}
    

\begin{table}[h]
\centering

\caption{Sim-to-Real Action Transfer Performance over 10 Trials}
\label{tab:real_world}
\begin{tabular}{lccc}
\hline
\textbf{Metric} & \textbf{Surface} & \textbf{Occluded} & \textbf{Perimeter} \\
\hline
Number of removals &1 &6.67 & 5.25\\
Objects moved &0 &0.5 & 0.25\\

\hline
\end{tabular}
\end{table}
The hardware validation was performed on three clutter types, with five trials on each type: 
\begin{itemize}
    \item \textbf{Surface Retrieval}: Evaluating robustness to direct target retrieval when the target is on the periphery of the clutter.
    \item \textbf{Multi Layer Occlusion}: Evaluating performance for cluttered stacks with multiple objects occluding the target.
    \item \textbf{Perimeter Confinement}: Evaluating performance in less dense clutter but with confined target objects.
\end{itemize}

Each scenario mentioned above demands different removal strategies, where simple heuristics like removing by decreasing stack height or target proximity would fail to generalise across all cases. Performance is evaluated via average removal and average number of objects moved, where a disturbance $>5$cm is considered a movement,  as shown in Table ~\ref{tab:real_world}. A supplementary video added with this work demonstrates human-like behaviour of the trained system. Fig. \ref{fig:rws}--\ref{fig:cd} show recorded logs for a single roll out in the real world.






\section{Conclusion and Future Work}

This work presented ClutterNav, a lightweight, interpretable framework for target retrieval in dense clutter using cost-based reinforcement learning and gradient-guided planning. The proposed approach ensures robustness and interpretability by providing a smooth, gradient-based importance measure and adapting to scene complexity with spatial re-weighting. The validation tests demonstrated that it achieves significant reduction in disturbances and removal steps compared to baselines while demonstrating human-like strategies in both simulation and hardware trials. An interesting direction to explore would be to use the same paradigm for learning with priors while including other objectives like Information Gain using inpainting based methods \cite{mapex} for generating better policies.




\begin{thebibliography}{00}
\bibliographystyle{IEEEtran}
\bibitem{binpick} M. Fujita, Y. Domae, A. Noda, G. A. Garcia Ricardez, T. Nagatani, A. Zeng,
S. Song, A. Rodriguez, A. Causo, I. M. Chen and T. Ogasawara , “What are the important technologies for bin picking? Technology analysis of robots in competitions based on a set of performance metrics,” Advanced Robotics, vol. 34, no. 7–8, 2020
\bibitem{old} C. Martinez, R. Boca, B. Zhang, H. Chen, and S. Nidamarthi, “Automated bin picking system for randomly located industrial parts,” in IEEE Conference on Technologies for Practical Robot Applications, TePRA, 2015. doi: 10.1109/TePRA.2015.7219656.
\bibitem{tamp} Y. Zhu, R. Mottaghi, E. Kolve, J. Lim, A. Gupta, L. Fei-Fei, and A. Farhadi, "Robotic Picking in Dense Clutter via Domain-Invariant Learning," in IEEE ICRA, 2019.
\bibitem{nam}C. Nam, J. Lee, S. Hun Cheong, B. Y. Cho, and C. H. Kim, “Fast and resilient manipulation planning for target retrieval in clutter,” in Proceedings - IEEE International Conference on Robotics and Automation, 2020. doi: 10.1109/ICRA40945.2020.9196652.
\bibitem{rlpick1} S. Cheong et al., “Obstacle rearrangement for robotic manipulation in clutter using a deep Q-network,” Intelligent Service Robotics, vol. 14, no. 4, 2021, doi: 10.1007/s11370-021-00377-4.
\bibitem{rlpick2} Y. Yang, H. Liang, and C. Choi, “A Deep Learning Approach to Grasping the Invisible,” IEEE Robotics and Automation Letters, vol. 5, no. 2, 2020, doi: 10.1109/LRA.2020.2970622.
\bibitem{ycb} B. Calli, A. Walsman, A. Singh, S. Srinivasa, P. Abbeel and A. M. Dollar, "Benchmarking in Manipulation Research: Using the Yale-CMU-Berkeley Object and Model Set," in IEEE Robotics and Automation Magazine, vol. 22, no. 3, pp. 36-52, Sept. 2015, doi: 10.1109/MRA.2015.2448951.
\bibitem{quickgrasp} N.S. Ravie, K. Vasan, A. Thondiyath, and B. Sebastian, "QuickGrasp: Lightweight Antipodal Grasp Planning with Point Clouds," arXiv preprint arXiv:2504.19716, 2025.

\bibitem{gpd}A. ten Pas, M. Gualtieri, K. Saenko, and R. Platt, “Grasp Pose Detection in Point Clouds,” International Journal of Robotics Research, vol. 36, no. 13–14, 2017, doi: 10.1177/0278364917735594.
\bibitem{SAM} Kirillov, A., Mintun, E., Ravi, N., Mao, H., Rolland, C., Gustafson, L., Xiao, T., Whitehead, S., Berg, A. C., Lo, W.-Y., and Dollár, P. “Segment Anything,” Proceedings of the IEEE/CVF International Conference on Computer Vision (ICCV), pp. 4015–4026, 2023.
\bibitem{pn++} C. R. Qi, H. Su, K. Mo, and L. J. Guibas, “PointNet: Deep learning on point sets for 3D classification and segmentation,” in Proceedings - 30th IEEE Conference on Computer Vision and Pattern Recognition, CVPR 2017, 2017. doi: 10.1109/CVPR.2017.16.

\bibitem{sac} T. Haarnoja, A. Zhou, P. Abbeel, and S. Levine, “Soft actor-critic: Off-policy maximum entropy deep reinforcement learning with a stochastic actor,” in 35th International Conference on Machine Learning, ICML 2018, 2018.

\bibitem{dgmvp} H. Ren, M. Yang, and S. Velipasalar, "3D Domain Generalization via Multiple Views of Point Clouds," arXiv preprint arXiv:2504.12456, 2024.

\bibitem{partGen} X. Wei, X. Gu, and J. Sun, "Learning Generalizable Part-based Feature Representation for 3D Point Clouds," in Advances in Neural Information Processing Systems (NeurIPS), vol. 35, pp. 12507-12519, 2022.

\bibitem{ig} M. Sundararajan, A. Taly, and Q. Yan, “Axiomatic attribution for deep networks,” in 34th International Conference on Machine Learning, ICML 2017, 2017.
\bibitem{jiang2024transic} Y. Jiang, C. Wang, R. Zhang, J. Wu, and L. Fei-Fei, “TRANSIC: Sim-to-Real Policy Transfer by Learning from Online Correction,” arXiv preprint arXiv:2405.10315, 2024.
\bibitem{mapex} C. Ho, S. Kim, B. Moon, A. Parandekar, N. Harutyunyan, C. Wang, K. Sycara, G. Best, and S. Scherer, "Mapex: Indoor structure exploration with probabilistic information gain from global map predictions," in 2025 IEEE International Conference on Robotics and Automation (ICRA), pp. 13074-13080, May 2025.


  
  
\end{thebibliography}
\end{document}